\documentclass{article} %
\usepackage{iclr2026_conference,times}

\usepackage{amsmath,amsfonts,bm}

\def\eqref#1{equation~\ref{#1}}

\def\1{\bm{1}}

\def\rvs{{\mathbf{s}}}

\DeclareMathAlphabet{\mathsfit}{\encodingdefault}{\sfdefault}{m}{sl}
\SetMathAlphabet{\mathsfit}{bold}{\encodingdefault}{\sfdefault}{bx}{n}

\def\sV{{\mathbb{V}}}

\newcommand{\E}{\mathbb{E}}

\newcommand{\Var}{\mathrm{Var}}

\DeclareMathOperator*{\argmax}{argmax}
\DeclareMathOperator*{\argmin}{argmin}

\usepackage[colorlinks=true,linkcolor=spearmint,citecolor=Orchid, urlcolor=pastel-red]{hyperref}
\usepackage{url}
\usepackage{notation}

\usepackage[table, dvipsnames]{xcolor}
\usepackage{xspace}
\usepackage{graphicx}
\usepackage[linesnumbered,ruled,vlined]{algorithm2e}

\usepackage{amsthm}
\usepackage{dsfont}
\usepackage{thmtools,thm-restate}
\usepackage[noabbrev,nameinlink,capitalize]{cleveref}

\SetCommentSty{mycommentstyle}

\usepackage{mathtools}
\DeclarePairedDelimiterX{\infdivx}[2]{(}{)}{%
  #1\;\delimsize\|\;#2%
}
\newcommand{\infdiv}{D_{\textrm{KL}}\infdivx}

\usepackage{multirow}
\usepackage{adjustbox}
\usepackage{booktabs}
\usepackage{wrapfig}

\usepackage{pgfplots}
\usepackage[table, dvipsnames]{xcolor}
\pgfplotsset{compat=1.18}
\usetikzlibrary{pgfplots.groupplots}

\definecolor{sanae0}{rgb}{0.9312692223325372, 0.8201921796082118, 0.7971480974663592}
\definecolor{sanae1}{rgb}{0.8559578605899612, 0.6418993116910497, 0.6754191211563135}
\definecolor{sanae2}{rgb}{0.739734329496642, 0.4765280683170713, 0.5959617419736206}
\definecolor{sanae3}{rgb}{0.57916573903086, 0.33934576125314425, 0.5219003947563425}
\definecolor{sanae4}{rgb}{0.37894937987024996, 0.2224702044652721, 0.41140014301575434}
\definecolor{sanae5}{rgb}{0.1750865648952205, 0.11840023306916837, 0.24215989137836502}
\definecolor{pastel-red}{HTML}{FF6961}
\definecolor{spearmint}{HTML}{4EB6B0}
\definecolor{teal}{HTML}{12486B}
\definecolor{cyan}{HTML}{9CE8F1}
\definecolor{violett}{HTML}{cdb4db}

\definecolor{darkblue}{rgb}{0, 0, 0.5}

\title{Entropy-Aligned Decoding of LMs \\for Better Writing and Reasoning}

\author{Kareem Ahmed \\
Department of Computer Science\\
University of California, Irvine\\
\texttt{ahmedky@uci.edu} \\
\And
Sameer Singh \\
Department of Computer Science\\
University of California, Irvine\\
\texttt{sameer@uci.edu} \\
}

\iclrfinalcopy %
\begin{document}

\maketitle

\begin{abstract}
Language models (LMs) are trained on billions of tokens in an attempt
to recover the true language distribution.
Still, vanilla random sampling from LMs yields low quality generations.
\emph{Decoding algorithms} attempt to restrict the LM distribution to a
set of high-probability continuations, but rely on greedy heuristics that
introduce myopic distortions, yielding sentences that are homogeneous,
repetitive and incoherent.
In this paper, we introduce \ours, a hyperparameter-free decoding approach
that incorporates the entropy of future trajectories into LM decoding.
\ours explicitly regulates the amount of uncertainty expressed at every step of generation, aligning the sampling distribution's entropy to the aleatoric (data) uncertainty.
Through \textsc{Entropy-Aware Lazy Gumbel-Max} sampling, \ours manages to be exact,
while also being efficient, requiring only a sublinear number of entropy evaluations
per step.
Unlike current baselines, \ours yields sampling distributions that are empirically
well-aligned with the entropy of the underlying data distribution.
Across creative writing and summarization tasks, \ours consistently improves
\textsc{LM-as-judge} preference win-rates over widely used decoding strategies.
These preference gains are complemented by automatic metrics, showing that \ours produces more diverse generations and more faithful summaries.
We also evaluate \ours on mathematical reasoning, where it outperforms all baselines.
\end{abstract}

\section{Introduction}
Despite the unprecedented capabilities of LMs at modeling natural language,
generating diverse, non-repetitive, and coherent text remains elusive.
Stochastic sampling tends to produce low quality text due
to sampling the unreliable tails of the distribution.\footnote{A limitation of softmax is that the resulting probability distribution always has full support.}
Consequently, many \emph{decoding algorithms} have been proposed that either statically~\citep{topk} or dynamically~\citep{topp, beamsearch, minp, etasampling, typicaldecoding}
intervene on the next-token distribution, in an effort to truncate, or restrict, the LM
distribution to a \emph{nucleus set} of the most promising sentences. 
Unfortunately, these heuristics are often ad hoc, based only on empirical
observations.
Furthermore, these ad hoc heuristics are often implemented \emph{greedily} at every
step of the generation, yielding a \emph{myopic} rendition of the desired target 
distribution.

Instinctively, one might be tempted to associate with high-quality \emph{human-like}
generations a high probability under the LM distribution.
Paradoxically, that does not seem to be the case.
In fact, empirical studies have shown~\citep{topp, zhang2021trading, meister2022paradox}
the quality of a generation to exhibits a peculiar relationship with regards to its
probability under the LM distribution: there is indeed a positive correlation between
the probability prescribed by a LM to a generation and its perceived quality \emph{up
to an inflection point} after which the perceived quality of the generation negatively
correlates with its probability.
This empirical observation offers a heuristic explanation for the phenomenon whereby
lower probability generations returned by stochastic decoding approaches appear
to outperform text generated using probability-maximizing approaches such as beam search~\citep{lowerre1976harpy, reddy1977speech, sutskever2014sequence, bahdanau2014neural}.

Information theory offers a principled account of why naive LM sampling fails.
The negative log-probability of an event corresponds to its \emph{surprisal};
under the asymptotic equipartition property (AEP)~\citep{shannon1948}, samples
from a distribution are overwhelmingly likely to fall in its typical set, with
probability mass concentrated near the entropy.
In language, typicality is intuitive: highly predictable strings convey little
information, while excessively surprising ones tend toward incoherence.
Stochastic autoregressive sampling from an LM is unbiased~\citep{koller2009probabilistic}\footnote{Each step samples exactly from the model’s conditional distribution.}, and so one might expect it to recover the true typical set.
However, even a small divergence between the LM and the true distribution can cause
an unbounded mismatch in entropy, and hence in typicality~\citep{braverman2020calibration}.
This distortion is further aggravated by mode collapse effects observed in finetuning~\citep{omahony2024attributing}, compounding the departure from
natural language behavior.

In this work, we introduce \ours, a hyperparameter-free decoding method that incorporates the entropy of future trajectories into language model decoding.
By correcting the entropy mismatch between the learned model and the true distribution, \ours overcomes the distortions of greedy step-wise heuristics.
Our approach combines lazy Gumbel-Max sampling~\citep{Hazan2013, Maddison2014, Mussmann2017FastAI} with admissible entropy upper bounds to efficiently prune candidate tokens, yielding an exact yet tractable algorithm for sampling from the entropy-aligned next-token distribution.
Since entropy is generally intractable to compute~\citep{Ahmed22nesyentropy, Valiant1979a, Valiant1979b}, and naïve Monte Carlo estimates suffer from high variance, we further introduce a Rao–Blackwellized estimator that achieves substantially lower variance and improved sample efficiency.
We evaluate \ours on AlpacaEval Creative Writing~\citep{alpacaeval}, GSM8K~\citep{Cobbe2021TrainingVT}, and CNN/DailyMail~\citep{nallapati-etal-2016-abstractive}, spanning creative writing, reasoning, and summarization.
Across benchmarks, \ours consistently outperforms top-$p$~\citep{topp}, top-$k$~\citep{topk}, and min-$p$~\citep{minp}, producing diverse yet
coherent generations.

\textbf{Contributions\;\;\;}
In summary, we introduce \ours, a decoding approach that calibrates, or \emph{aligns}, the entropy of the LM with that of the true language distribution.
This is achieved by reweighting the model's next-token distribution with a quantity proportional to the entropy of the distribution over future trajectories.
Leveraging lazy Gumbel-Max sampling together with admissible entropy upper bounds, \ours efficiently prunes candidate tokens while retaining exactness.
Monte Carlo estimation of entropy is prohibitively high-variance, so we develop
a low-variance, sample-efficient, Rao–Blackwellized entropy estimator.
Our experiments across creative writing, reasoning, and summarization benchmarks show \ours consistently improves the coherence and diversity of generations.

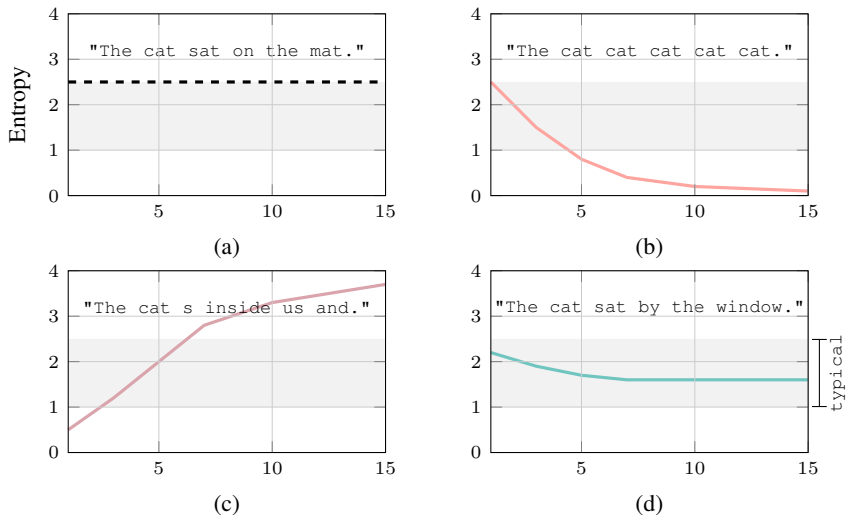
\begin{figure}[t!]
\centering
\begin{tikzpicture}
  \begin{groupplot}[
    group style={group size=2 by 2, horizontal sep=1.4cm},
    width=5.8cm, height=4cm,
    ylabel={Entropy},
    xmin=1, xmax=15, ymin=0, ymax=4,
    grid=both,
    major grid style={line width=0.3pt, draw=gray!40},
    minor grid style={line width=0.2pt, draw=gray!20},
    axis on top,
    enlarge x limits=false,
    enlarge y limits=false,
    clip=false,
    tick label style={font=\scriptsize},
    label style={font=\small}
  ]

  \nextgroupplot
    \addplot [fill=gray!10, draw=none, domain=1:15] {2.5} \closedcycle;
    \addplot [fill=white,   draw=none, domain=1:15] {1.0} \closedcycle;
    \addplot [black, dashed, very thick] coordinates {(1,2.5)(15,2.5)};
    \node[font=\scriptsize, align=center] at (axis cs:8.0,3.2)
      {\texttt{"{}The cat sat on the mat."{}}};
    \node[font=\small, anchor=north] at (rel axis cs:0.5,-0.18) {(a)};

  \nextgroupplot[ylabel={}]
    \addplot [fill=gray!10, draw=none, domain=1:15] {2.5} \closedcycle;
    \addplot [fill=white,   draw=none, domain=1:15] {1.0} \closedcycle;
    \addplot [pastel-red!60, very thick] table {
      1 2.5
      3 1.5
      5 0.8
      7 0.4
      10 0.2
      15 0.1
    };
    \node[font=\scriptsize, align=center] at (axis cs:8.0,3.2)
      {\texttt{"{}The cat cat cat cat cat."{}}};
    \node[font=\small, anchor=north] at (rel axis cs:0.5,-0.18) {(b)};

  \nextgroupplot[ylabel={}]
    \addplot [fill=gray!10, draw=none, domain=1:15] {2.5} \closedcycle;
    \addplot [fill=white,   draw=none, domain=1:15] {1.0} \closedcycle;
    \addplot [sanae1!100, very thick] table {
      1 0.5
      3 1.2
      5 2.0
      7 2.8
      10 3.3
      15 3.7
    };
    \node[font=\scriptsize, align=center] at (axis cs:8.0,3.2)
        {\texttt{"{}The cat s inside us and."{}}};
    \node[font=\small, anchor=north] at (rel axis cs:0.5,-0.18) {(c)};

  \nextgroupplot[ylabel={}]
    \addplot [fill=gray!10, draw=none, domain=1:15] {2.5} \closedcycle;
    \addplot [fill=white,   draw=none, domain=1:15] {1.0} \closedcycle;
    \addplot [spearmint!80, very thick] table {
      1 2.2
      3 1.9
      5 1.7
      7 1.6
      10 1.6
      15 1.6
    };
    \node[font=\scriptsize, align=center] at (axis cs:8.0,3.2)
      {\texttt{"{}The cat sat by the window."{}}};
    \node[font=\small, anchor=north] at (rel axis cs:0.5,-0.18) {(d)};

    \draw[|-|, , xshift=0.3cm]
  ($(axis cs:15,1.0)+(0.5,0)$) --
  ($(axis cs:15,2.5)+(0.5,0)$)
  node[midway, sloped, below=3pt, inner sep=0.5pt]{\scriptsize \texttt{typical}};

  \end{groupplot}

\end{tikzpicture}
\caption{%
\textbf{Entropy trajectories during text generation conditioned on \texttt{"The cat"} prompt.}
Figure (a) shows a reference sample with entropy values staying in the ``typical'' band (gray region). 
Figure (b) illustrates degeneration into repetition, where entropy collapses below the typical range. 
Figure (c) shows gibberish, where entropy rises well above the typical range. 
Figure (d) demonstrates \ours decoding, which maintains entropy close to the typical band and yields coherent text.
}
\end{figure}

\section{Related Works}
Sampling methods are crucial in controlling the quality and diversity of text generated by LLMs.
The choice of sampling strategy directly affects the balance between creativity and coherence, which is critical in many generative tasks.
In this section, we review existing approaches and their
limitations.

\textbf{Greedy Decoding and Beam Search\;\;\;} Greedy decoding and beam search ~\citep{lowerre1976harpy, reddy1977speech, sutskever2014sequence, bahdanau2014neural} are deterministic decoding
strategies that select the token with the highest probability at each step \citep{freitag2017beam}.
While these methods ensure high-probability token selection, they often lead to repetitive and generic
text due to their lack of diversity. Beam search also incurs a significant runtime performance penalty.

\begin{wrapfigure}{r}{0.5\textwidth}
\vspace{-1.5em}
    \centering
    \includegraphics[width=\linewidth, trim={0 0 0 0}, clip]{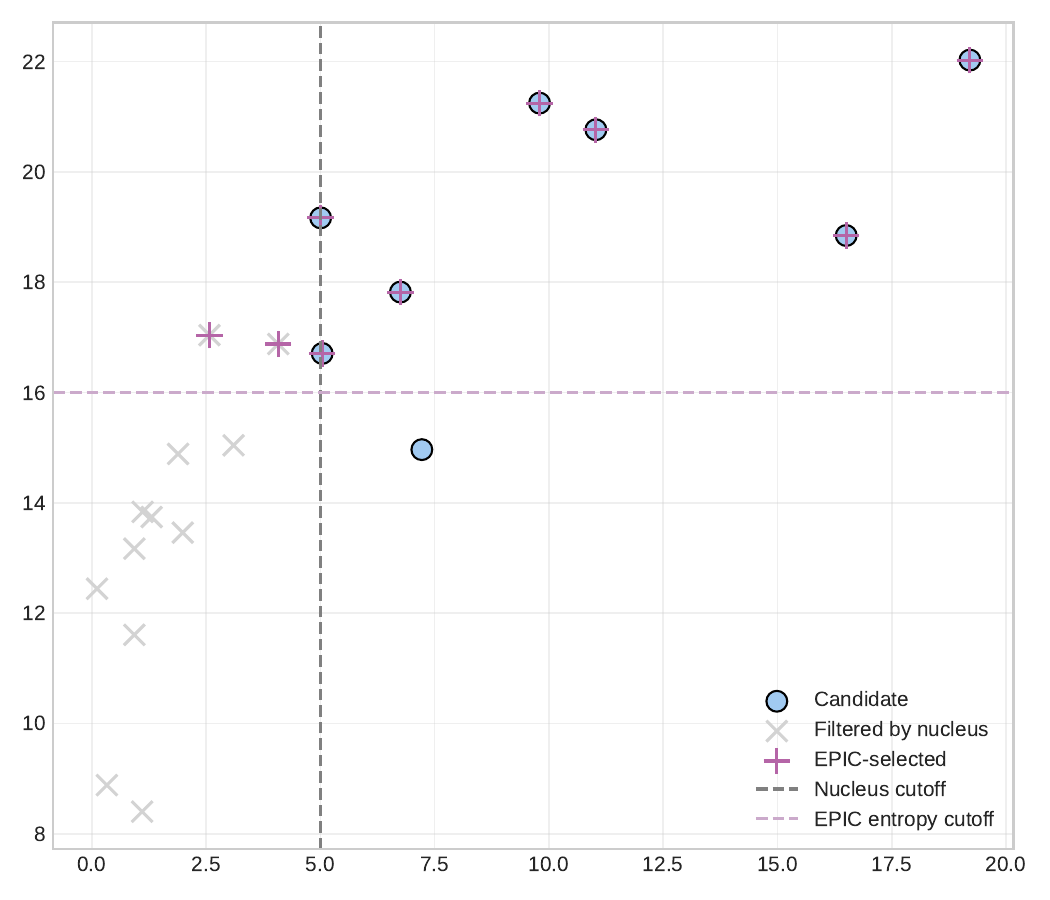}
\vspace{-2em}
\caption{
\textbf{Interaction of \ours and truncation.}
Points denote candidate tokens, represented by probability ($\rightarrow$) and lookahead entropy ($\uparrow$).
Gray crosses denote truncated tokens.
Tokens considered by truncation algorithms are denoted using blue circles, while \ours-candidate tokens are denoted by violet pluses.
The vertical dashed line denotes the probability cutoff, and the horizontal dashed line denotes the entropy cutoff.
\ours favors tokens that balance both probability and future uncertainty, avoiding degenerate low-entropy repetitions or incoherent high-entropy expansions.
}
\vspace{-2em}
\end{wrapfigure}
\textbf{Stochastic Sampling Methods\;\;\;} Stochastic sampling methods aim to inject diversity into the
generated text by introducing randomness in token selection. Temperature scaling adjusts the
distribution’s sharpness, balancing diversity and coherence \citep{ackley1985boltzmann}; however, higher
temperatures often lead to incoherent and nonsensical results, limiting its applicability. Top-$k$
sampling selects from the top $k$ most probable tokens, ensuring that only high-probability tokens
are considered \citep{fan2018hierarchical}. While it offers a simple way to prevent unlikely tokens from being
sampled, it does not adapt dynamically to varying confidence levels across different contexts.
Top-$p$ sampling, also known as nucleus sampling, restricts the token pool to those whose cumulative
probability exceeds a predefined threshold $p$ \citep{holtzman2020nucleus}. This method effectively balances
quality and diversity by focusing on the ``nucleus'' of high-probability tokens and dynamically adapts
to different contexts. However, at higher temperatures, top-$p$ sampling can still allow low-probability
tokens into the sampling pool, leading to incoherent outputs. This trade-off between creativity and
coherence at high temperatures is a key limitation that min-$p$ sampling has aimed to address.

\textbf{Entropy-Based Methods\;\;\;} Recent work has introduced methods such as entropy-dependent
truncation ($\eta$-sampling) and mirostat sampling, which attempt to dynamically adjust the sampling
pool based on the entropy of the token distribution \citep{hewitt2022entropy, basu2021mirostat}.
While entropy- and uncertainty-based approaches show promise in improving text quality, the local
heuristics they rely on are often ad hoc, and their parameters tend to be unintuitive and difficult to tune.

\textbf{Controllable Generation Approaches\;\;\;}
Our work bears resemblance to prior works that bias LM samples towards satisfying
syntactic~\citep[\textit{inter alia}]{ahmed2023a, Ahmed2025controllable, Honghua2024Adaptable, Geh2025tokenization,guidance,outlines, vanKrieken2025neurosymbolic_diffusion_models, koo2024}, or semantic constraints~\citep[\textit{inter alia}]{ahmed2025semantic_probabilistic_control, YidouWengICML25, Zhao-et-al-2024-twist-functions-2024-ICML, loula2025syntactic-2025-ICLR, yang-klein-2021-fudge,Dathathri2020Plug,liu-etal-2021-dexperts,Beurer-Kellner-ICML2024-DOMINO,kumar-etal-2022-gradient,qin2022cold,pmlr-v235-du24a, pynadath2025controlled-dab}, steering LM generations through adjusting next-token probabilities.
Conversely, our setting differs in that the constraint is distributional rather than token-level, or even sequence-level: we require that samples be drawn from a restricted subset of the model’s support that satisfies certain distributional properties, rather than merely biasing next-token probabilities toward syntactic or semantic criteria.

\section{The Typical Set Paradox in LMs}
Autoregressive sampling from a language model produces sequences by drawing each token
in turn from the model’s conditional distribution.
This ensures an unbiased generation process whereby the sentences are sampled
according to the model's joint distribution.
Importantly, such samples overwhelmingly come from \emph{the typical set} of
the distribution, rather than from the single most likely sequence.
Roughly speaking, although there is a staggering amount of samples that might
be generated by a random process, the one actually produced is almost surely
from a loosely defined set of samples that all have approximately the same
chance of being the one actually realized.
And despite individual samples whose probability dominates that of any single
outcome in this set, the vast number of realizations in the set all but guarantees
the the outcome will come from such set.

To make this more intuitive, consider flipping a biased coin ($p=0.7$), $n$ times.
The most likely sequence is the one where all flips turn up heads, and has probability $0.7^n$.
By contrast, the typical set consists of sequences with roughly $0.7n$ heads and $0.3n$ tails.
Each of these sequences has probability roughly $(0.54)^n$, which is smaller than $0.7^n$.
However, there are about $\binom{n}{0.7n} \approx 2^{nH(0.7)}$ such sequences.
As a result, nearly all of the probability mass lies in this exponentially large
typical set, not on the single mode sequence of all heads, even though the latter
is individually more likely.
Concretely, for $n=20$, the probability of the all heads sequences is a meager $0.08$
compared to the collective probability of flipping 13, 14, or 15 heads, which accounts
for over $53\%$ of the mass.

\textbf{This raises a paradox:} \emph{how can we assert that autoregressive sampling draws
from the typical set, yet simultaneously argue for a decoding method to bias generation toward
that very same set?}

\subsection{Typical Set Misalignment under Model Miscalibration}
\label{subsec:typical_misalignment}
Concretely, let us denote an LM generation of arbitrary length $T$ by $\y_{1:T} \coloneqq
\left[y_1 y_2 \dots y_T \right]$, where $y_i$ is the instantiation of random variable $Y_i$
and takes values from a fixed vocabulary $\sV = \{1, \dots, V\}$.

For a distribution $\p$, we denote by $H(\p)$ its \emph{entropy}, defined as $H(\p) = \E_{\y_{1:T}
\sim \p}\left[- \log \p(\y_{1:T})\right]$.
Additionally, for a distribution $\q$, we denote by $H(\p, \q)$ the \emph{cross-entropy} between
the distributions $\p$ and $\q$, defined as $H(\p, \q) = - \E_{\y_{1:T} \sim \p}[\log \q(\y_{1:T})]$, quantifying the average number of bits required to encode samples from $\p$ using a code optimized for $\q$.
We can also re-write the \emph{cross-entropy} as $H(\p, \q) = H(\p) + \infdiv{p}{q}$, where $\infdiv{p}{q}$ denotes the \emph{KL-divergence} between $\p$ and $\q$, defined as $\infdiv{\p}{\q} = \E_{\y_{1:T} \sim \p}\left[\log \p(\y_{1:T}) - \log \q(\y_{1:T})\right]$, and quantifies the expected extra number of bits needed to encode samples from distribution $\p$ when using a code optimized for $\q$.

Note that for a \emph{calibrated} language model, we would hope that $H(\p, \q) \approx H(\q)$ \ie it's uncertainty regarding its own generations matches the uncertainty it exhibits on data drawn from the true distribution.
Furthermore, throughout the paper we will make the assumption that the model is \emph{accurate}
\begin{equation}\label{eqn:kl-epsilon}
\infdiv{p}{q} = H(\p, \q) - H(\p) < T \cdot \varepsilon,
\end{equation}
\ie we assume \emph{uncertainty} due to imperfect knowledge of $p$ can be made arbitrarily small, so that the model's uncertainty primarily reflects the (irreducible) \emph{aleatoric uncertainty} intrinsic to the data.

\paragraph{Prologue.} Next, we will show that even under the aforementioned assumption that the epistemic uncertainty due to imperfect knowledge of the distribution can be made arbitrarily small, the mismatch in the entropy between the model and the true underlying distribution can grow unboundedly.

We start by redefining our learned model $q$ as $\q\coloneqq(1-\varepsilon) \mathbin{\cdot} \q + \varepsilon \cdot \mathcal{U}$, where $\mathcal{U}$ denotes the uniform distribution over $\y_{1:T}$.
This is to ensure that $\q(\y_{1:T}) > 0$ for all $\y_{1:T}$, as is standard in softmax-based LMs.
Now consider the \emph{bounded} function $f = -\log \q$.
If the learned model $\q$ is accurate, in the sense of~\cref{eqn:kl-epsilon}, one might hope that the expected value of $f$ under the true distribution $p$ would be close to the expected value under $\q$, \ie $\mu_\p(f) \approx \mu_\q(f)$. However, as we will show next, the model $\q$ might be \emph{$\varepsilon$-accurate} and still suffer from entropy miscalibration under long generations.

\begin{restatable}{lemma}{miscalibration}[Entropy Miscalibration~\citep{braverman2020calibration}]\label{lemma:miscalibration}
Suppose the bound in~\cref{eqn:kl-epsilon} holds. The calibration error between the true distribution $\p$ and the learned model $\q$ is bounded~as
\begin{equation}
\label{eqn:miscalibration_bound}
    \lvert H(\p, \q) - H(\q)\rvert \leq \sqrt{2\varepsilon(T+1)}(T\log M + \log(1/\varepsilon)).
\end{equation}
\end{restatable}
The last inequality is the \emph{entropy miscalibration bound}, and it
clearly shows that, in the worst case, even a small cross
entropy may provide little control over the generations under the learned model. In fact,
for $\varepsilon = O(\frac{1}{T})$, which we may hope is an accurate model,
the bound turns out to be vacuous.
\subsection{Entropy Calibration via Entropy-Aligned Decoding}

To address the aforementioned \emph{entropy miscalibration}, we propose explicitly \emph{calibrating the model distribution to the entropy functional}, the idea being to define a family of reweighted distributions
\begin{equation}\label{eq:tilted_distribution}
\q_{t, \alpha}(y_t \mid \y_{<t}) \propto \q_{t}(y_t \mid \y_{<t}) \exp\left(-\alpha \, H_{t:t+k}(y_t)\right),\footnote{This recovers \emph{global entropy calibration}, \ie $\q_{\alpha}(\y) \propto \me^{\alpha - \log \p(\y)} \cdot \p(\y) = \p(\y)^{1+\alpha} $ reducing to \emph{global temperature scaling}, decomposing token-wise as $\p_{t, \alpha}(y_t \mid \y_{<t}) \propto \p_{t}(y_t \mid \y_{<t})^{1+\alpha} \cdot \E_{\y_{> t}}[\p(\y_{> t} \mid \y_{<t})^{1+\alpha}]$.}
\end{equation}
where $H_{t:t+k}(y_t) \coloneqq \E_{\Y_{t+1:t+k}\sim\q(\cdot \mid \y_{\leq t}))}[H(\Y_{t+1:t+k} \mid \y_{\leq t})]$ denotes the \emph{$k$-step lookahead entropy} associated with prefix $\y_{<t}$, and $\alpha$ is a learnable calibration parameter. 
Intuitively, the \emph{entropy-aligned} distribution in~\cref{eq:tilted_distribution} biases the sampling process according to the estimated lookahead entropy of a prefix: sequences expected to devolve into low-entropy repetitions or high-entropy gibberish are down weighted, whereas those expected to maintain a calibrated entropy are amplified.

We now show that the \emph{entropy-aligned} distribution in~\cref{eq:tilted_distribution} is such that the model’s $k$-step lookahead entropy is calibrated to the true distribution without degrading the model's accuracy. Furthermore, $\alpha^{*}$ is the unique minimizer of $H(\p, \q_{\alpha})$, and can be efficiently computed using bisection.

\begin{restatable}{lemma}{EPIC}[Entropy-Aligned Decoding Lowers CE and Calibrates Entropy]\label{lemma:EAD}
Let $\p$ be the true distribution, and $\q$ be the learned model.
For horizon $k$, prefix $\y_{<t}$, and candidate token $y_t$, let
\begin{equation}
    H_{t:t+k}(y_t) \coloneqq \E_{\Y_{t+1:t+k}\sim\q(\cdot \mid \y_{\leq t})}[H(\Y_{t+1:t+k} \mid \y_{\leq t})].
\end{equation}
Define the \emph{entropy-aligned} distribution $\q_{t, \alpha}(y_t \mid \y_{<t})$ as
\begin{equation}
\q_{t, \alpha}(y_t \mid \y_{<t}) \propto \q_{t}(y_t \mid \y_{<t}) \exp\left(-\alpha \, H_{t:t+k}(y_t)\right).
\end{equation}
Let
\begin{equation}
    \mu_{\mathcal{D}} \coloneqq \sum_{t=1}^{T} \E_{\y<t \sim \p, y_t \sim \mathcal{D}(\cdot \mid \y_{<t})}\left[H_{t:t+k}(y_t)\right].
\end{equation}
Then there exists $\alpha^{*} \in \mathbb{R}$ such that
\begin{enumerate}
    \item $\alpha^{*} = \argmin_{\alpha} H(\p, \q_{\alpha})$ {\color{Cerulean}[$\alpha^{*}$ minimizes CE of the true and entropy-aligned distributions]}
    \item $\mu_{\q_{\alpha^{*}}} = \mu_{\p}$ {\color{Cerulean}[The model's entropy is calibrated to the true entropy]}
    \item $H(\p, \q_{\alpha^{*}}) \leq H(\p, \q)$ {\color{Cerulean}[Entropy Aligned decoding does not worsen accuracy]}
\end{enumerate}
\end{restatable}
\cref{lemma:EAD} suggests a simple algorithm.
Find the $\alpha^*$ that minimizes the cross entropy on a held-out set, or a larger model's generations prompted on the target task.
Then, at every timestep, for every token, estimate the lookahead entropy, reweigh the logits by the estimated lookahead entropies, renormalize and sample.
The reader might observe that such a simple algorithm is not a very practical one: \emph{it requires that we perform $O(\lvert\mathcal{V}\rvert)$ forward passes of the LM at every time step.}
In what follows, \emph{we show how we can compute the cross-entropy and draw exact samples from the entropy-aligned distribution while evaluating the lookahead entropy for only a handful of candidate tokens.}

\section{Entropy-Aware Lazy Gumbel-Max for Sublinear Sampling}
\label{sec:lazy-gumbel-entropy}

\paragraph{Background: the Gumbel-Max trick.}
For any given timestep, let $\{\tilde{y}_i\}_{i=1}^{\mathcal{\lvert V\rvert}}$ denote the \emph{log unnormalized} scores, or \emph{logits}, parameterizing the LM's next-token distribution.
One means of sampling from the above distribution is to add \iid \emph{Gumbel noise} to each of the logits and take the $\argmax$:
\begin{equation}
y^\star = \argmax_i \{\tilde{y}_i + G_i\} \sim \mathrm{Categorical}\left(\frac{e^{y_i}}{\sum_j e^{y_j}}\right) \text{\!, where }
G_i \sim \mathrm{Gumbel}(0,1).
\label{eq:gumbel-max}
\end{equation}
The resulting sample $y$ is precisely distributed according to the \emph{normalized} target distribution~\citep{Hazan2013, Maddison2014}. We will call $\tilde{y}_i+G_i$ the \emph{perturbed scores} or \emph{perturbed logits}.

\paragraph{Lazy (sublinear) Gumbel-Max.}
A naive application of Gumbel-Max samples \emph{all} $\mathcal{\lvert V\rvert}$ Gumbels.
A key observation of \emph{lazy Gumbel-Max}~\citep{Mussmann2017FastAI}
is that \emph{a maximizer of the perturbed logits either posses a large
$\tilde{y}_i$, or realizes an unusually large $G_i$}.
Consequently, we can avoid instantiating all Gumbels
by (a) selecting the top-$k$ scores $\mathcal{S}=\mathtt{TopK}(\{\tilde{y}_i\},k)$ and
instantiating their Gumbels, and (b) \emph{lazily} drawing \emph{only} Gumbels exceeding the gap to the front-runner in the~tail.\footnote{Sample the expected number of Gumbels  $m$ exceeding the gap. Then, uniformly sample a subset of size $m$ from the set $\mathcal{\lvert V \rvert}\setminus\mathcal{S}$. Assign each element in the subset a Gumbel drawn conditionally greater than the bound.}

\paragraph{Entropy-Aware Lazy Gumbel-Max.}
Applying Gumbel-Max to~\cref{eq:tilted_distribution} reduces to taking
\begin{equation}
 y^\star = \argmax_{i\in\mathcal{V}}
\left\{\tilde{y}_i + G_i - \alpha \cdot H_{t:t+k}(y_i)\right\}
\label{eq:perturbed-entropy-score},
\end{equation}
requiring we compute the lookahead entropy $H_{t:t+k}(y_i)$ for each token $y_i \in \mathcal{V}$, which is infeasible.

\paragraph{Bounding $H_{t:t+k}(\cdot)$ for lazy selection.}
Taking inspiration from \emph{lazy Gumbel-Max}, we can \emph{bound} $H_{t:t+k}(\cdot)$, and therefore the Entropy-Aligned Perturbed Logits (\earl) in~\cref{eq:perturbed-entropy-score}.
We can then consider \emph{only} the candidates likely to overtake the front-runner.
More precisely, for a token $y_i$,
\begin{equation}
\tilde{y}_i+G_i-\alpha \cdot \mathtt{H_t^U(k)}
\le
s_i
\le
\tilde{y}_i+G_i-\alpha \cdot \mathtt{H_t^L(k)},
\label{eq:score-bounds}
\end{equation}
\begin{wrapfigure}{r}{0.57\textwidth}
\vspace{-1.5em}
    \centering
    \includegraphics[scale=0.3]{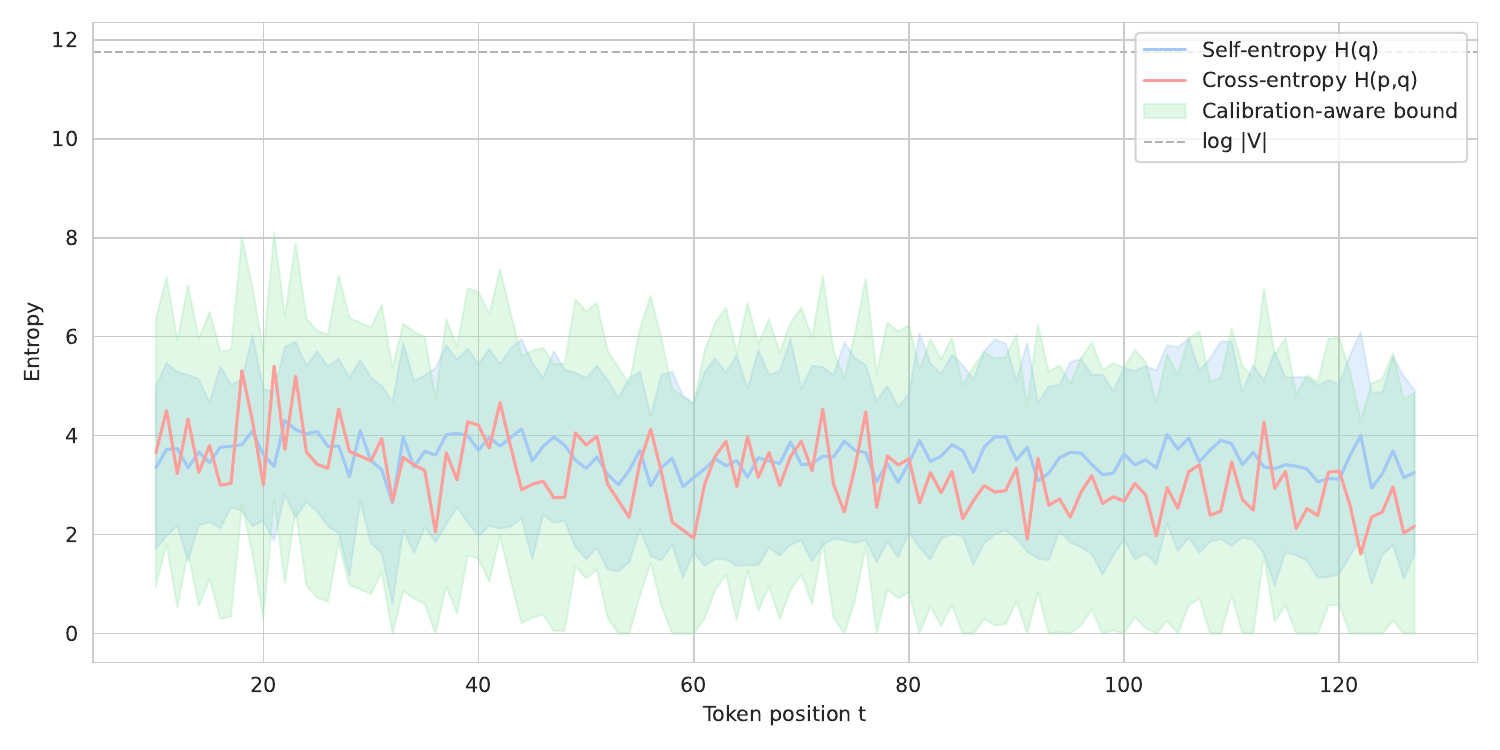}
\vspace{-1em}
\caption{
\textbf{Tighter Entropy Bounds.} Making use of \cref{lemma:miscalibration} in tandem with the empirical model error lends to much tighter entropy bounds than theoretical.}
\label{fig:empirical_bounds}
\vspace{-2em}
\end{wrapfigure}
we can \emph{loosely} bound the entropy from below by $0$ and from above by 
$T \cdot \log \mathcal{\lvert V \rvert}$, the theoretical minimum and maximum,
respectively, such that the \earl score of every token $y_i$ is \emph{admissibly}
bounded.
We can, however, further tighten the above bounds by making use of the result in~\cref{lemma:miscalibration} in tandem with the model error to arrive at much tighter bounds (see~\cref{fig:empirical_bounds}).
We will now show that in light of these bounds, we only need a handful of new entropy evaluations to ascertain the actual winner.

\begin{restatable}{lemma}{expectedevaluations}[Expected Entropy Evaluations under Bounded Correction]
Let $\{\tilde{y}_i\}_{i=1}^{\mathcal{\lvert V\rvert}}$ denote the \emph{log unnormalized} scores, or \emph{logits}, parameterizing the LM's next-token distribution.
Let $\{G_i\}$ denote \iid standard Gumbel random variables.
let $z_i$ denote the centered perturbed-logits, $z_i = (l_i + g_i) - \log \sum_j \me^{l_j}$.
Assume we have an $\hat{H}_i$ such that $\lvert H_{t:t+k}(y_i) - \hat{H}_i\rvert \leq C_t$. Let $w \coloneqq \lvert\alpha\rvert \cdot C_t$.
Define plug-in scores $u_i\coloneqq z_i - \alpha \hat{H}_i$,
then it is the case that every true score $s_i$ lies in the
interval $s_i \in [u_i - w , u_i + w]$.
and we have that the expected number of entropy evaluations is $\E\left[N_{\text{eval}}\right] = \me^{2w}$.
\end{restatable}
Instantiating the above lemma with an average per-step error of $\varepsilon = 10^{-3}$, and $\alpha =0.2$, the maximum we have encountered across domains in our experiments, yields only $4$ expected evaluations.

\textbf{Entropy-Aligned Decoding (\ours)\ }
We are now ready to give a full treatment of our \ours procedure.
To reiterate, our goal is to sample the next token using Gumbel-max
from the \emph{entropy-tilted} next-token distribution given in~\cref
{eq:perturbed-entropy-score}, while \emph{avoiding an expensive lookahead
over the full vocabulary}.
To that end, we start by perturbing the next-token logits with Gumbel
noise on~\cref{line:epic_gumbels}.
Next, on~\cref{line:epic_start_eliminate}-\cref{line:epic_end_eliminate}
candidates that are $\lvert\alpha\rvert \cdot (\mathtt{H_t^U(k)} - 
\mathtt{H_t^L(k)})$ apart from $\mathtt{frontrunner}$ are eliminated from
the race: such candidates can't overtake the $\mathtt{frontrunner}$ even
if the $\mathtt{frontrunner}$ only attains the lower bound while the
candidates attain the upper bound.
If a single candidate remains, it is declared a winner, terminating our
procedure (\cref{line:single_candidate}).
Otherwise, we proceed by tightening our bounds (\cref{line:start_tighten}-\cref{line:end_tighten}), computing the one-step lookahead entropy and
upper-bounding the remaining \earl scores by adding the most favorable
value consistent with the bounds, given by $\max\{- \alpha \cdot 
\mathtt{H_t^U(k-1)}, - \alpha \cdot \mathtt{H_t^L(k-1)}\}$ on~\cref{line:upb_scores}.
This retains an admissible upper bound that accounts for the sign of $\alpha$.\footnote{Depending on the domain, $\alpha$ can assume either positive or negative values. Positive $\alpha$ biases decoding towards low-entropy, predictable continuations, while negative $\alpha$ favors high-entropy, diverse continuations.}
Finally, candidates are examined block-wise (\cref{line:uncover_block}),
in descending order of their upperbounded \earl scores, computing the
lookahead entropies only as needed on a per-block basis.
Evaluation stops once a candidate’s \earl score exceeds the next-best
upperbounded \earl score, at which point it is guaranteed to be optimal.
We adaptively compute the lookahead horizon\footnote{We give the definition for the $\mathtt{GetRolloutHorizn}$ function in the appendix.} (\cref{line:horizn}) by
estimating the minimum number of rollout steps required for a candidate’s
optimistic score to exceed the current threshold.
The rollout budget is set to the smallest horizon for which the remaining entropy
term could close the observed score gap, capped by a global maximum.
This strategy allocates lookahead computations only where it can affect the
decision, yielding sizable efficiency gains in practice.
Our full \ours procedure is given in~\cref{alg:EPIC}.

\textbf{Correctness\ }
Note that our \ours procedure detailed above maintains for each candidate index $i$
bounds $\mathrm{LB}_i\le s_i \le \mathrm{UB}_i$ with $\mathrm{LB}_i$ non-decreasing
and $\mathrm{UB}_i$ non-increasing as bounds are refined.
Furthermore, the stopping condition guarantees the returned $y^\star$ satisfies
$s_{y^\star} \ge \mathrm{UB}_j$ for all $j\neq y^\star$, and that no unvisited tail
index $j$ can have  $H_{t:t+k}(y_i)$ large enough to make $s_j \ge s_{y^\star}$.
Therefore the algorithm returns $\argmax_i s_i$, \ie an exact sample
from~\cref{eq:perturbed-entropy-score}, and therefore,~\cref{eq:tilted_distribution}.

\begin{algorithm}[t !]
\caption{Entropy-Aligned Decoding Logits Processor}
\label{alg:EPIC}
\KwIn{Sequence $\mathtt{input\_ids}$, next-token $\mathtt{scores}$, calibration parameter $\alpha$, and lookahead $k$}
\KwOut{A token $y_i$ sampled from the entropy-aligned $\p_{\alpha}(Y_i|\y_{<i}) \propto  \p(Y_i|\y_{<i})\cdot \me^{-\alpha \cdot H(\Y_{i:i+k})}$}
\vspace{0.2cm}
\tcc{Get non-zero candidate tokens and corresponding scores}
$\mathtt{candidates} = \mathtt{scores.isfinite().nonzero(as\_tuple=True)}$\;
$\mathtt{scores = scores[scores.isfinite()]}$\;
\vspace{0.2cm}
\tcc{Perturb the logits using Gumbel noise}
$\mathtt{gumbel\_logits} = \mathtt{-\log(-\log(rand\_like(scores)))} + \mathtt{scores}$\;\label{line:epic_gumbels}
\vspace{0.2cm}
\tcc{Eliminate candidates separated by $\alpha \times \mathtt{H_t^U}$ from $\mathtt{frontrunner}$ as they can't win even if the $\mathtt{frontrunner}$ only attains the lowerbound entropy $\mathtt{H_t^L}$ while they attain the upperbound $\mathtt{H_t^U}$}
$\mathtt{frontrunner} = \mathtt{max(gumbel\_logits)}$\;\label{line:epic_start_eliminate}
$\mathtt{mask} = \mathtt{gumbel\_logits}  > \mathtt{frontrunner} - \lvert\alpha\rvert \cdot (\mathtt{H_t^U(k)} - \mathtt{H_t^L(k)})$\;
$\mathtt{gumbel\_logits}, \mathtt{candidates} = \mathtt{gumbel\_logits}[\mathtt{mask}], \mathtt{candidates}[\mathtt{mask}]$\; \label{line:epic_end_eliminate}
\vspace{0.2cm}

\tcc{Only a single candidate left after pruning}
\If{$\mathtt{len(candidates)} == 0$}{\Return $\mathtt{candidates}[0]$}\label{line:single_candidate}
\vspace{0.2cm}

\tcc{Tighten bounds by computing $\mathtt{step\_entropy}$ corresponding to 1-step lookahead entropy}
$\mathtt{lookaheads} = \mathtt{concat}(\mathtt{input\_ids}, \mathtt{candidates})$\; \label{line:start_tighten}
$\mathtt{step\_entropy} = \mathtt{Categorical}(\mathtt{model}(\mathtt{lookaheads})).\mathtt{entropy()}$\;

$\mathtt{upb\_scores} = \mathtt{gumbel\_logits} - \alpha \cdot \mathtt{step\_entropy} + \max\{- \alpha \cdot \mathtt{H_t^U(k-1)}, - \alpha \cdot \mathtt{H_t^L(k-1)}\}$\;\label{line:upb_scores}
$\mathtt{sorted\_scores}, \mathtt{sorted\_candidates} = 
   \mathtt{sort}(\mathtt{upb\_scores}, \mathtt{candidates})$\;\label{line:end_tighten}
\vspace{0.2cm}
\tcc{Uncover entropy-weighted scores for the top candidates}
\For{each block $\mathtt{B}$ in $\mathtt{sorted\_candidates}$}{ \label{line:uncover_block}
    $\mathtt{lookaheads} = \mathtt{concat}(\mathtt{input\_ids}, \mathtt{B})$\;
    $\mathtt{threshold} = \mathtt{sorted\_scores}[\mathtt{B}+1] \;\; \text{if } \mathtt{B}+1 < \text{len}(\mathtt{sorted\_scores}) \;\; \text{else } -\infty$\;
    $\mathtt{horizn} = \mathtt{GetRolloutHorizn}(\mathtt{sorted\_scores}[\mathtt{B}], \mathtt{step\_entropy}[\mathtt{B}], \alpha, k, \mathtt{threshold})$\;\label{line:horizn}
    $\mathtt{lookahead\_entropies} = \mathtt{rb\_lookahead\_entropy}(\mathtt{model}, \mathtt{lookaheads}, \mathtt{horizn})$\;

    $\mathtt{candidate\_scores} = \mathtt{sorted\_scores}[\mathtt{B}] + \alpha \cdot \mathtt{lookahead\_entropies}$\;

    \If{$\max(\mathtt{candidate\_scores}) > \mathtt{threshold}$}{
        \Return $\mathtt{sorted\_candidates[argmax(candidate\_scores)]}$
    }
}
\end{algorithm}

\begin{figure}
    \centering
    \includegraphics[ width=\linewidth]{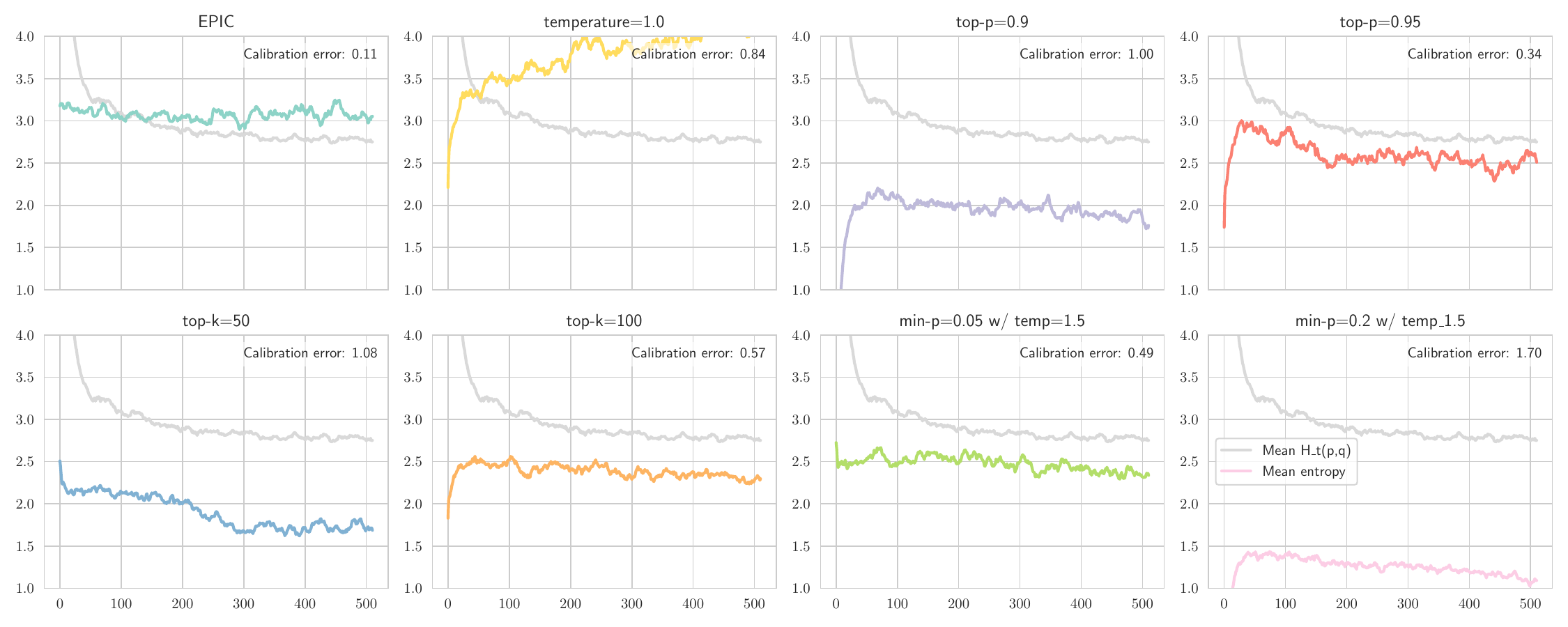}
    \caption{\textbf{Calibration of Decoding Methods to Model Cross-Entropy.}
We plot the mean conditional entropy of generated sequences as a function of the reference cross-entropy, along with the resulting calibration error for each decoding strategy. \ours closely tracks the target cross-entropy across the full range, yielding substantially lower calibration error and a markedly smoother trajectory than competing methods. In contrast, standard heuristics such as top-$p$, top-$k$, and min-$p$ exhibit systematic bias and high variance, leading to persistent miscalibration despite aggressive tuning.}
    \label{fig:calibration}
\end{figure}

\section{Experiments}\label{sec:exps}
In this section, we explore the efficacy of our \ours decoding strategy.
First off, we qualitatively ascertain that, using a humble lookahead horizon
of merely $k=4$, \ours is able to track the target entropy achieving almost
perfect calibration.
Next, we show that such qualitative merits are coupled with meaningful improvements
in generation quality, as measured both using an \textsc{LM-as-judge}~\citep{zheng2023judging}
as well as automatic quality and diversity metrics.
More concretely, we evaluate $\ours$ on three language generation tasks: abstractive summarization, create story generation, and mathematical reasoning.
Our hope in curating such a diverse set of benchmarks is to establish \ours's ability to preserve the fidelity of the articles to be summarized, generate diverse yet coherent
stores in response to creative prompts, and improve the reasoning abilities of LMs.
We assess \ours's performance with respect to several other stochastic decoding strategies: nucleus sampling, top-$k$ sampling, typical decoding, and temperature sampling.\footnote{Temperature sampling is defined as ancestral sampling after renormalization with an annealing term $\tau$}
For writing tasks, we report the win-rate (\textsc{WR}), the length-controlled win-rate (\textsc{LC-WR}).
Additionally, for abstractive summarization we report \textsc{BERTScore-F1}~\citep{zhang2020bertscore} to measure semantic coverage, and for creative writing we report \textsc{Self-BLEU}~\citep{montahaei2019SelfBleu} to measure the generation lexical diversity.
Lastly, for mathematical reasoning we report the accuracy of predicting the correct answer.

\subsection{Calibration to Cross-Entropy}

We evaluate how well different decoding approaches preserve the entropy profile of the base model by measuring their calibration to cross-entropy.
For each method, we generate sequences across a range of operating points and compute the
mean entropy rate of the resulting samples.
We then compare this quantity to the corresponding reference cross-entropy, reporting both
the full trajectory and an aggregate calibration error.
As shown in~\cref{fig:calibration}, \ours closely tracks the target cross-entropy across
the entire range.
Its entropy trajectory is smooth and stable, resulting in substantially lower calibration error
than all competing approaches.
In contrast, standard decoding heuristics such as top-\(p\), top-\(k\), and min-\(p\) exhibit
systematic bias away from the target entropy, leading to persistent miscalibration even when
their hyperparameters are tuned.
This behavior highlights a key distinction between \ours and prior decoding methods:
While conventional approaches adjust next-token probabilities locally, \ours constrains the
induced sequence-level distribution.
As a result, \ours maintains global entropy alignment, yielding smoother entropy dynamics and improved calibration.

\subsection{Quantitative Experiments}\label{sec:setup}
\textbf{Implementation and Data\ } We use the HuggingFace framework~\citep{transformers}, employing their implementations of top-$p$, top-$k$, min-$p$, temperature sampling,
and typical decoding. 
For story generation, we evaluate on the \textsc{WritingPrompts} dataset~\citep{fan2018hierarchical}.
For abstractive summarization, we evalaute on the \textsc{CNN/DailyMail}
dataset~\citep{nallapati-etal-2016-abstractive}.
In a preliminary hyperparameter sweep we determine the optimal hyperparameter
values on a small validation set using the \textsc{LM-as-judge} metric.
For mathematical reasoning, we evaluate on the GSM8K dataset~\citep{cobbe2021gsm8k}.
The value of $\alpha$ in \ours is determined using the procedure outlined in~\cref{sec:estimating_alpha}.
All reported metrics are computed on the respective test sets, or subsets thereof.

\subsubsection*{Evaluation metrics\ }
For writing tasks, we evaluate generation quality using \textsc{LM-as-judge}~\citep{zheng2023judging}.
We fix a \emph{reference} decoding strategy\textemdash the best-performing variant of 
min-$p$\textemdash and compare all other methods against it in pairwise evaluations.
For each prompt, the LM judge is presented with a pair of generations, one from the
reference method and the other from the candidate method.
The LM is then asked to ascertain whether the reference wins, the candidate wins, or there 
was a tie.
All evaluations are conducted anonymously, with generations presented in randomized
order to avoid positional or method-specific bias.
We repeat this procedure across $10$ random seeds using the same underlying set of prompts
and generations, reporting the win rate (\textsc{wr}) and the length-controlled win
rate (\textsc{lc-wr}), thereby controlling for generation length, in an effort to eliminate
length bias.

Furthermore, to assess lexical diversity in creative writing, we report 
\textsc{Self-BLEU}~\citep{montahaei2019SelfBleu} scores, where lower values indicate greater
diversity among generated outputs.
\textsc{Self-BLEU} measures the average BLEU score of each generation against the remaining
samples, and thus captures mode collapse or excessive similarity across generations.
We treat \textsc{Self-BLEU} as a complementary diagnostic rather than a standalone quality
metric
seeing as diversity taken alone is not indicative of generation quality.
Furthermore, we report \textsc{BERTScore-F1 Score}~\citep{zhang2020bertscore} on abstractive summarization to measure semantic coverage with respect to the source article.

\begin{table}[t!]
\centering
\setlength{\tabcolsep}{4pt}
\caption{
\textbf{Comparison of the Different Decoding Approaches on the \textsc{WritingPrompts} and \textsc{CNN/DailyMail} Datasets.} The win rate (\textsc{wr}) and length-controlled win rate are computed using ChatGPT-$5$, against min-$p$, and are averaged across $10$ seeds. \textsc{BLEU} denotes the
\textsc{Self-BLEU} score, measuring the diversity of each generation against the remaining samples. \textsc{F1} denotes the \textsc{BERTScore-F1 Score}, capturing how well the generated output preserves the source content.\\}

\begin{tabular}{lcccccc}
\toprule
 & \multicolumn{3}{c}{\textsc{WritingPrompts}} 
 & \multicolumn{3}{c}{\textsc{CNN/DailyMail}} \\
\cmidrule(lr){2-4} \cmidrule(lr){5-7}
\textbf{Decoding Approach} 
 & \textsc{wr} ($\uparrow$) & \textsc{lc-wr} ($\uparrow$) & \textsc{BLEU} ($\downarrow$)
 & \textsc{wr} ($\uparrow$) & \textsc{lc-wr} ($\uparrow$) & \textsc{F1} ($\uparrow$)\\
\midrule
Top-$k$ $({k}=50)$         
 & 54\% & 54\% & 1.84
 & 49\% & 49\% & 0.16 \\
Top-$p$ $({p}=0.9)$          
 & 51\% & 51\% & 1.92
 & 52\% & 52\% & 0.18 \\
Temperature $({\tau}=1.5)$              
 & 0\% & 0\% & \textbf{0.69}
 & 2\% & 2\% & $-$0.24 \\
Typical $({\tau}=0.95)$                 
 & 43\% & 43\% & 1.80
 & 49\% & 49\% & 0.17 \\
 Min-$p$ $(p = 0.2, \tau = 1.5)$                 
 & $-$ & $-$ & 2.12
 & $-$ & $-$ & 0.17 \\
\midrule
\textbf{\ours (ours)}    
 & \textbf{58\%} & \textbf{58\%} & 1.55
 & \textbf{56\%} & \textbf{55\%} & \textbf{0.19} \\
\bottomrule
\end{tabular}
\label{fig:wp_cnn_full}
\end{table}

\begin{wraptable}[17]{r}{0.45\textwidth}
\centering
\vspace{-1em}
\caption{
Decoding approaches evaluated on a random subset of the GSM8k dataset.\\
}
\begin{tabular}{lc}
\toprule
\textbf{Decoding Method} 
 & \textbf{Accuracy} ($\uparrow$) \\
\midrule
Greedy          
 & $79.27$ \% \\
Top-$k$ $({k}=50)$          
 & $71.95$ \% \\
Top-$k$ $({k}=100)$        
 & $68.29$ \% \\
Top-$p$ $(p = 0.9)$           
 & $74.39$ \% \\
 Top-$p$ $(p = 0.95)$           
 & $74.39$ \% \\
Typical $(\tau = 0.2)$            
 & $73.17$ \% \\
Typical $(\tau = 0.95)$              
 & $74.39$\% \\
 min-$p$ $(p = 0.05, \tau = 1.5)$                
 & $60.98$\% \\
 min-$p$ $(p = 0.2, \tau = 1.5)$                 
 & $75.61$\% \\
 \midrule
\textbf{\ours (ours)}    
 & \textbf{$\mathbf{85.37}$\%} \\
\bottomrule
\end{tabular}
\vspace{-2.5em}
\label{tab:GSM8K}
\end{wraptable}
\par\smallskip
\noindent
\vspace{-2.5em}
\subsubsection*{Results}

We first evaluate \ours on open-ended text generation tasks, including creative writing on the \textsc{WritingPrompts} dataset and abstractive summarization on \textsc{CNN/DailyMail}.
Our results are shown in~\cref{fig:wp_cnn_full}.
Across both tasks, \ours consistently outperforms competing decoding strategies when evaluated using an \textsc{LM-as-judge}.
On \textsc{WritingPrompts}, \ours achieves the highest preference win-rate among all methods, substantially exceeding standard decoding heuristics such as top-\(p\), top-\(k\), and min-\(p\).
Importantly, these gains persist under length-controlled evaluation, indicating that improvements are not driven by verbosity but reflect higher perceived quality of the generated stories.
These preference-based results are further corroborated by diversity metrics.
\ours attains a markedly lower Self-BLEU score than almost all baseline methods, in fact matching the \textsc{Self-BLEU} achieved by the human reference generations.
This suggests that \ours produces a more diverse set of outputs, avoiding the mode collapse and repetitive patterns commonly observed in entropy-reducing decoding strategies.
Note that while the temperature baseline achieves a much lower \textsc{Self-BLEU}, indicative of higher diversity, it achieves a 0\% win-rate compared to Min-$p$.
In fact, upon inspecting the outputs of the temperature baselines, generations almost
always devolved into incoherent gibberish.
This goes to the point we made earlier: measures of diversity are only meaningful
when considered in tandem with measures of generation quality.
On \textsc{CNN/DailyMail}, \ours again achieves the highest preference win-rate under the \textsc{LM-as-judge} evaluation.
In addition, \ours attains the best \textsc{BERTScore-F1 score} among the evaluated methods, indicating improved semantic coverage of the source articles.
While the absolute differences in \textsc{F1} are modest, they are consistent across runs and align with the preference-based judgments, suggesting that \ours better preserves salient content without over-constraining generation.
We note that we make use of the official \textsc{BERTScore} implementation\footnote{\url{https://github.com/Tiiiger/bert_score}}, with $\mathtt{rescale\_with\_baseline=True}$.
Consequently, the BERTScore is normalized relative to the expected similarity between unrelated sentences for the underlying encoder. 
As a result, scores are centered around zero, with negative values indicating worse-than-chance semantic alignment. While this leads to lower absolute values, it yields a calibrated metric that better reflects semantic fidelity and avoids inflated scores due to embedding bias.
We also evaluate \ours on \textsc{GSM8K}, shown in~\cref{tab:GSM8K}, to assess its performance on structured mathematical reasoning tasks.
\ours outperforms the strongest baseline approach by an almost $5\%$ absolute accuracy, demonstrating that its advantages extend beyond open-ended generation to settings requiring precise multi-step reasoning.

\subsubsection*{Conclusion}
We introduced \ours, an entropy-aware decoding approach for LMs.
\ours steers language model generations by explicitly aligning the sampling distribution to a target entropy profile, capturing the irreducible (aleatoric) uncertainty of plausible continuations.
We show that, unlike existing decoding baselines, \ours yields sampling distributions that are empirically well-aligned with the entropy of the underlying data distribution.
Across creative writing and summarization tasks, \ours consistently improves
\textsc{LM-as-judge} preference win-rates over widely used decoding strategies:
top-$k$, top-$p$, min-$p$ typical decoding, and temperature scaling.
Importantly, these gains persist under length-controlled evaluation, indicating that \ours’s improvements are not driven by superficial verbosity effects but by genuinely higher-quality generations.
These preference gains are complemented by automatic metrics, which show that \ours produces more diverse generations in creative settings and more faithful summaries in summarization tasks.
We further evaluate \ours on mathematical reasoning, where it outperforms considered baselines.
Through \textsc{Entropy-Aware Lazy Gumbel-Max}, \ours manages to be exact for any given horizon $k$, while also being efficient and lightweight.

\subsubsection*{Acknowledgments}
This work is supported in part by the DARPA ANSR program FA8750-23-2-0004 as well as an NSF CAREER award number IIS-2046873. The conclusions presented within this research paper are of the authors and do not reflect the official policy or position of DARPA or the U.S. Government.

\bibliography{iclr2026_conference}
\bibliographystyle{iclr2026_conference}

\appendix
\section{Proofs}
\miscalibration*
\begin{proof}
Pinsker's Inequality tells us that for any bounded function $f$ with $\lVert f \rVert_{\infty} < B$, we have that
\begin{equation}
    \lvert \E_\p[f] - \E_\q[f]\rvert \leq B \sqrt{2\infdiv{p}{q}}.
\end{equation}
We will start by deriving a bound on $\infdiv{p}{q}$, followed by deriving a bound on $B$.

Recall our re-definition of $\q$ in~\cref{subsec:typical_misalignment} as a mixture, where, explicating the dependence on $\varepsilon$
\begin{equation}
    \qeps(\y_{1:T}) = (1 - \varepsilon) \cdot \q(\y_{1:T}) + \varepsilon \cdot M^{-T} \geq (1 - \varepsilon) \cdot \q(\y_{1:T}),
\end{equation}
and therefore,
\begin{equation}
    \frac{1}{\qeps(\y_{1:T})} \leq  \frac{1}{(1 - \varepsilon) \cdot \q(\y_{1:T})}.
\end{equation}
Taking the logarithm, which is a monotonically increasing function, we get
\begin{align}
    \log \frac{1}{\qeps(\y_{1:T})} &\leq  \log \frac{1}{(1 - \varepsilon) \cdot \q(\y_{1:T})}\\
    & = \log \frac{1}{q(\y_{1:T})} + \log \frac{1}{1 - \varepsilon}.
\end{align}
Relaxing $\log \frac{1}{1-\varepsilon}$ using a simple bound valid for $\varepsilon \in (0, \frac{1}{2}]$, we have that
\begin{equation}
\log \frac{1}{1-\varepsilon} = - \log (1-\varepsilon) \leq 2\varepsilon,
\end{equation}
and therefore,
\begin{equation}
\label{eqn:logq_bound}
\boxed{
    \log \frac{1}{\qeps(\y_{1:T})} \leq \log \frac{1}{\q(\y_{1:T})} + 2\varepsilon.}
\end{equation}
Expanding the expression for the KL-divergence and substituting the bound in~\cref{eqn:logq_bound}
\begin{equation}
    \infdiv{\p}{\qeps} = \E_{\p}[\log \frac{\p}{\qeps}] = \E_{\p}[\log \frac{1}{\qeps}] - \E_{\p}[\log \frac{1}{\p}] = \E_{\p}[\log \frac{1}{\q}] - \E_{\p}[\log \frac{1}{\p}] + 2\varepsilon.
\end{equation}
Substituting the bound in~\cref{eqn:kl-epsilon}, we get our bound on the KL-divergence term,
\begin{equation}
\label{eqn:kl_bound}
\boxed{
    \infdiv{\p}{\qeps} \leq T\varepsilon + 2\varepsilon.}
\end{equation}
Lastly, deriving a value for $B$, we bound the maximum value of $-\log\qeps(\y_{1:T})$ as
\begin{equation}
\label{eqn:B_bound}
\boxed{
    -\log\qeps(\y_{1:T}) \leq -\log(\varepsilon M^{-T}) = \log \frac{M^T}{\varepsilon} = T\log M + \log \frac{1}{\varepsilon}.}
\end{equation}
Substituting~\cref{eqn:kl_bound} and~\cref{eqn:B_bound} into~\cref{eqn:miscalibration_bound}, we get our desired result.
\end{proof}
\EPIC*
\begin{proof}
We will write $S_t(y_t) \coloneqq H_{t:t+k}(y_t)$ for brevity.
We will also write the token-wise entropy-aligned distribution, making explicit the dependence on the partition function $Z_t(\alpha; \y_{<t})$, as
\begin{equation}
\q_{t, \alpha}(y_t \mid \y_{<t}) = \frac{\q_{t}(y_t \mid \y_{<t}) \exp\left(-\alpha S_t(y_t)\right)}{Z_t(\alpha; \y_{<t})}, \text{ where } Z_t(\alpha; \y_{<t}) = \sum_{y_t} \q_t(y_t \mid \y_{<t})\exp{(-\alpha S_t(y_t))}.
\end{equation}
Taking the derivative of the log-partition function $Z_t(\alpha; \y_{<t})$ \wrt $\alpha$,
we get
\begin{equation}\label{eq:log-partition-derivative}
    \frac{\partial}{\partial \alpha} \log Z_t(\alpha; \y_{<t}) = -\E_{y_t \sim \q_{t, \alpha}(\cdot \mid \y_{<t})}[S_t(y_t)].
\end{equation}
Moreover, the second derivative
\begin{equation}
    \frac{\partial^2}{\partial^2 \alpha} \log Z_t(\alpha; \y_{<t}) = \Var_{y_t \sim \q_{t, \alpha}(\cdot \mid \y_{<t})}[S_t(y_t)] \geq 0,
\end{equation}
and therefore, $\log Z_t(\alpha; \y_{<t})$ is convex in $\alpha$.
We will also write the sequence-level cross-entropy
\begin{equation}
    F(\alpha) \coloneqq H(\p, \q_\alpha) = \sum_{t=1}^T \E_{\y_{<t} \sim \p} \E_{y_{t} \sim \p(\cdot \mid \y_{<t})}[- \log \q_{t, \alpha}(y_t\mid \y_{<t})]
\end{equation}
and differentiating \wrt $\alpha$
\begin{equation}\label{eq:cross-entropy-derivative}
    F'(\alpha) = \sum_{t=1}^T  \E_{\y_{\leq t} \sim \p} [S_t(y_t)] + \sum_{t=1}^T  \E_{\y_{< t} \sim \p}[\frac{\partial}{\partial \alpha} \log Z_t(\alpha; \y_{<t})].
\end{equation}
Substituting~\cref{eq:log-partition-derivative} into~\cref{eq:cross-entropy-derivative}, we get
\begin{equation}
\label{eq:calibration}
     F'(\alpha) = \sum_{t=1}^T \E_{\y_{< t} \sim \p}[\E_{y_{t} \sim \p(\cdot \mid \y_{<t})}\left [S_t(y_t)] - \E_{y_t \sim \q_{t, \alpha}(\cdot \mid \y_{<t})}[S_t(y_t)]\right] = \mu_{\p} - \mu_{\q_\alpha}
\end{equation}
Since $F(\alpha)$ is a sum of a constant, an affine function, as well as the convex log-partition function, it follows that $F(\alpha)$ is convex.
Since $F$ is convex and continuous in $\alpha$, $\alpha^*$ is a minimizer of $F$.

Next, having showed $F'(\alpha) = \mu_{\p} - \mu_{\q_\alpha}$~\cref{eq:calibration}, and since $F'(\alpha) = 0$, we have $\mu_{\p} - \mu_{\q_{\alpha^*}}$, which establishes calibratio.
Lastly, since $\alpha^*$ minimizes $F(\alpha)$, then  $F(\alpha^*) = H(\p, \q_{\alpha^*}) \leq F(0) = H(\p, \q_{0}) = H(\p, \q)$, which proves that entropy-aligned decoding does not degrade accuracy.
\end{proof}
\expectedevaluations*
\begin{proof}
Let $M = \max_{i} z_i$. Then for any $y \in \mathbb{R}$, the probability of the maximum $M < y$ is given by
\begin{equation}
    \p(M \leq y) = \prod_{i=1}^n\p(z_i \leq y) = \prod_{i=1}^n \exp(-\me^{-(y - \log a_i)}) = \exp(-\me^{-y} \sum_i a_i) = \exp(-\me^{-y}),
\end{equation}
with $a_i = \frac{\me^{l_i}}{\sum_j \me^{l_j}}$ the probability of token $y_i$. Then, $M$ is Gumbel with density $f_M(y) = \me^{-y}\me^{-\me^{-y}}$.

We denote by $T$ the \emph{count of $z_i$} such that each $z_i$ is within a $w$-window of the maximum $M$, \ie 
\begin{equation}
    T = \#\{i: z_i > M - w\} = \sum_i \ind{z_i > M - w}.
\end{equation}
where we're interested in computing the value of $T$ on average. One observation is that, conditional on $M=y$, each non-max indicator $\ind{z_i > M - w}$ is an independent Bernoulli with probability
\begin{equation}
    \p_i(y) \coloneqq \p(z_i > y - w) = 1 - \exp(-a_i \me^w \me^{-y}),
\end{equation}
and consequently,
\begin{equation}
    \E[T|M=y] = \sum_i \cdot p_i(y).
\end{equation}
Averaging out all values of $M$ using the law of total expectation, we get
\begin{equation}
    \E[T] = \E_M[\E[T|M]] =  \sum_i \E[\p_i(M)] =  \sum_i \int p(y) f_M(y) dy
\end{equation}
Using the fact that if $M \sim \mathrm{Gumbel}(0,1)$ then $X\coloneqq\me^{-M} \sim \mathrm{Exp(1)}$, we have that for any given $z_i$
\begin{equation}
    \E[p(M)] = \E[1 - \me^{-a_i \me^w \me^{-M}}] = \E[1 - \me^{-a_i \me^w X}] = \int_0^\infty 1 - \me^{-a_i \me^w x} e^{-x} dx = \frac{a_i \me^w}{1 + a_i \me^w}.
\end{equation}
Therefore, 
\begin{equation}
    \E[T] = \sum_i \frac{a_i \me^w}{1 + a_i \me^w}.
\end{equation}
For any given token probability $a$, define
\begin{equation}
    f(a) = \frac{a\me^w}{1 + a\me^w}, \qquad a \in [0, 1]
\end{equation}
which is concave on the interval $(0, 1]$. Therefore, by Jensen's inequality, we have that
\begin{equation}
    \frac{1}{n} \sum_i f(a_i) \leq f(\frac{1}{n}\sum_i a_i) = f(\frac{1}{n}),
\end{equation}
and,
\begin{equation}
    \E[T] = \sum_i f(a_i) \leq n \cdot f(\frac{1}{n}) = \frac{n\me^w}{n + \me^w} < \me^w.
\end{equation}
That is, for any set of next-token probabilities $\{a_i\}_{i=1}^n$ it holds that $\E[T] < \me^w$.

Moving from a one-sided uncertainty band of width $w$, where only challengers may
gain, to a two-sided band of width $2w$, where the incumbent may decrease and
challengers may increase, enlarges the feasible window from $[-w,0]$ to $[-2w,0]$,
and since the expected number of near- maximal points scales as $\me^w$,
the candidate count increases from $\me^{w}-1$ to $me^{2w}-1$, concluding our proof.
\end{proof}

\section{Implementation Details}

\subsection{Efficiently Estimating $\alpha$}
\label{sec:estimating_alpha}

Recall from~\cref{lemma:EAD} that $\alpha^{*} = \argmin_{\alpha} 
H(\p, \q_{\alpha}) = \argmin_{\alpha} \E_p[-\log \q_\alpha(Y)]$.
Unfortunately, as written, computing the cross-entropy, and by extension
the minimizer of the cross-entropy is computationally intractable due to
the dependence on $\q_\alpha$.
In what follows, we will show how we can efficiently estimate $\alpha$ by
leveraging importance sampling while side stepping the above intractability.

We start by rewriting the cross entropy of the distribution $p$ \wrt the entropy-aligned distribution $\q_\alpha$
\begin{equation}
\mathcal{C}(\alpha) \coloneqq \E_p[-\log q_\alpha(Y)] = \E_p[-\log q(Y)] + \alpha \E_p[H(Y)] + \log Z(\alpha).
\label{eq:CE_app}
\end{equation}
Since only the last two terms depend on $\alpha$, by differentiating, we get
\begin{equation}
\mathcal{C}'(\alpha)
=
\E_p[H(Y)]
+
\frac{d}{d\alpha}\log Z(\alpha),
\end{equation}
where
\begin{equation}
\frac{d}{d\alpha}\log Z(\alpha)
=
-\E_{\q_\alpha}[H(Y)],
\end{equation}
and therefore,
\begin{equation}
\mathcal{C}'(\alpha)
=
\E_p[H(Y)]
-
\E_{\q_\alpha}[H(Y)].
\label{eq:gradCE_app}
\end{equation}

We therefore have that any minimizer $\alpha^\star$ of $\mathcal{C}(\alpha)$
satisfies the moment-matching condition
\begin{equation}
\E_p[H(Y)]
=
\E_{\q_{\alpha^\star}}[H(Y)].
\label{eq:moment_match_app}
\end{equation}
That is, the expected entropy under the entropy-aligned model matches the
true expected entropy.

Moreover,
\begin{equation}
\frac{d}{d\alpha}\E_{\q_\alpha}[H(Y)]
=
-\Var_{\q_\alpha}\!\bigl(H(Y)\bigr)
\;\le\;0,
\end{equation}
so $\E_{p_\alpha}[H]$ is nonincreasing in $\alpha$.
Furthermore, letting
\begin{equation}
\label{eq:moment_matching}
g(\alpha)
\;\coloneqq\;
\E_p[H(Y)] - \E_{p_\alpha}[H(Y)],
\end{equation}
we have that $g'(\alpha)\ge 0$. Therefore, the function $g$ is monotone nondecreasing in $\alpha$.
Consequently, the solution to the minimization problem $\alpha^*$ is unique, and can be efficiently found using bisection.

What remains is to show how to efficiently estimate the two expectations in~\cref{eq:moment_matching}:
\begin{itemize}
\item
$\mu_p \equiv \E_p[H(Y)]$, estimated from a held-out validation dataset
$\{y^{(m)}\}_{m=1}^M$ via
\begin{equation}
\widehat{\mu}_p
=
\frac{1}{M}\sum_{m=1}^M H(y^{(m)}), \text{ and;}
\end{equation}

\item
$\mu_\alpha \equiv \E_{\q_\alpha}[H(Y)]$, which admits the importance-sampling identity
\begin{equation}
\mu_\alpha
=
\frac{\E_q\!\left[ H(Y)\,e^{-\alpha H(Y)} \right]}
     {\E_q\!\left[ e^{-\alpha H(Y)} \right]},
\label{eq:IS_app}
\end{equation}
estimated using samples $\tilde{y}^{(i)}\sim q$ and importance weights $w_\alpha(\tilde{y})=e^{-\alpha H(\tilde{y})}$.
\end{itemize}

\subsection{Efficient Entropy Estimation via Rao-Blackwellization}

\paragraph{Naïve Monte Carlo estimator for $H_{t:t+k}(\cdot)$}
One means of estimating the entropy $H_{t:t+k}(\cdot)$ is to
simulate $K$ rollouts $\{\y^{(i)}_{t:t+k}\}_{i=1}^K$ of length $k$ from $\p$, and averaging their negative log-probabilities 
\begin{equation}
    \Hat{H}_{t:t+k}^{\mathrm{MC}}(y_t) = -\frac{1}{K}\sum_{i=1}^K p(\y_{t+1:t+k} \mid \y_{<t}).\end{equation}
While the above estimator is unbiased for large values of $K$, it suffers from very high variance.

\paragraph{Rao-Blackwellized estimator\!} By the chain rule of entropy, we can write the lookahead entropy as 
\begin{equation}
H\left(\Y_{t+1:t+k}\mid\y_{\le t}\right)
 = \sum_{j=1}^{k} \mathbb{E}\left[H\left(Y_{t+j}\mid \Y_{\le t+j-1}\right)\right],
\label{eq:chain}
\end{equation}
where the inner term $H(Y_{t+j}\mid Y_{\le t+j-1})$ is the \emph{entropy of a one-step predictive distribution} that the model exposes directly from its logits at the corresponding prefix.
\cref{eq:chain} therefore admits a low-variance \emph{Rao-Blackwellized} estimator that replaces the empirical entropy of joint rollouts by the \emph{exact} conditional entropies, averaged only over the (random) prefixes, which can be written as
\begin{equation}
\Hat{H}_{t:t+k}^{\mathrm{RB}}(y_t) = \sum_{j=1}^{k}\;
\frac{1}{K}\sum_{i=1}^{K}
H\left(Y_{t+j}\,\mid \y^{(i)}_{\leq t+j-1}\right)
\label{eq:rb-est}
\end{equation}
By the law of total variance,
\begin{equation}
\mathrm{Var}\left(\Hat{H}_{t:t+k}^{\mathrm{RB}}\right)
=
\mathrm{Var}\bigl( \mathbb{E}[\Hat{H}_{t:t+k}^{\mathrm{MC}} \mid \mathcal{F}] \bigr)
\le
\mathrm{Var}\left(\Hat{H}_{t:t+k}^{\mathrm{MC}}\right),
\quad
\mathcal{F}=\{p(Y_{t+j}\mid \Y^{(i)}_{\leq t+j-1})\}_{i,j},
\label{eq:variance-reduction}
\end{equation}
with strict inequality unless the conditional entropies are constant. Intuitively, \cref{eq:rb-est} collapses the
Monte Carlo noise due to \emph{which} token is drawn at step $t{+}j$ by conditioning on the prefix and using the \emph{analytic} entropy
of $p(Y_{t+j}\,\mid \y^{(i)}_{\leq t+j-1})$, for all $j$. In practice, we found $K \in \{2,4\}$ to yielding very low variance estimates of the entropy,
yielding a fast and stable entropy estimator.

\end{document}